\documentclass{article}

    \usepackage[preprint]{neurips_2021}

\usepackage[utf8]{inputenc} % allow utf-8 input
\usepackage[T1]{fontenc}    % use 8-bit T1 fonts
\usepackage{hyperref}       % hyperlinks
\usepackage{url}            % simple URL typesetting
\usepackage{booktabs}       % professional-quality tables
\usepackage{amsfonts}       % blackboard math symbols
\usepackage{nicefrac}       % compact symbols for 1/2, etc.
\usepackage{microtype}      % microtypography
\usepackage{xcolor}         % colors
\usepackage{amsmath}
\usepackage{microtype}
\usepackage{graphicx}
\usepackage{subfigure}
\usepackage{bm}
\usepackage{mathtools}
\usepackage{xspace}
\usepackage{enumitem}
\usepackage{bbm}
\usepackage{amsthm}
\usepackage{multirow}
\usepackage{cancel}
\usepackage{algorithm}
\usepackage{algorithmic}
\usepackage{caption}

\newif\ifcomm
\commtrue
\ifcomm
	\newcommand{\mycomm}[3]{{\footnotesize{{\color{#2} \textbf{[#1: #3]}}}}}
	\newcommand{\CRdel}[1]{\textcolor{red}{\sout{#1}}}
\else
    \newcommand{\mycomm}[3]{}
    \newcommand{\CRdel}[1]{}
\fi
\newcommand{\norm}[1]{\|#1\|}
\newcommand{\nth}{\textsuperscript{th}\xspace}

\DeclareMathOperator*{\argmax}{arg\,max}
\DeclareMathOperator{\mwp}{mwp}
\DeclareMathOperator{\topa}{top}
\DeclareMathOperator{\trunc}{trunc}

\newtheorem{theorem}{Theorem}[section]

\newtheorem{asm}{Assumption}
\newtheorem{definition}{Definition}

\title{Towards Federated Learning With Byzantine-Robust Client Weighting}

\author{
  Amit Portnoy\\
  Department of Computer Science\\
  Ben-Gurion University of the Negev\\
  \texttt{amitport@post.bgu.ac.il}\\
  \And
  Yoav Tirosh\\
  Department of Computer Science\\
  Ben-Gurion University of the Negev\\
  \texttt{yoavti@post.bgu.ac.il}\\
  \AND
  Danny Hendler\\
  Department of Computer Science\\
  Ben-Gurion University of the Negev\\
  \texttt{hendlerd@cs.bgu.ac.il}\\
}

\begin{document}

\maketitle

\begin{abstract}
  \emph{Federated Learning} (FL) is a distributed machine learning paradigm where data is distributed among clients who collaboratively train a model in a computation process coordinated by a central server. By assigning a weight to each client based on the proportion of data instances it possesses, the rate of convergence to an accurate joint model can be greatly accelerated. Some previous works studied FL in a Byzantine setting, in which a fraction of the clients may send arbitrary or even malicious information regarding their model. However, these works either ignore the issue of data unbalancedness altogether or assume that client weights are a priori known to the server, whereas, in practice, it is likely that weights will be reported to the server by the clients themselves and therefore cannot be relied upon. We address this issue for the first time by proposing a practical weight-truncation-based preprocessing method and demonstrating empirically that it is able to strike a good balance between model quality and Byzantine robustness. We also establish analytically that our method can be applied to a randomly selected sample of client weights.
\end{abstract}

\section{Introduction}\label{sec:intro}

\emph{Federated Learning} (FL) \citep{ konevcny2015federated, mcmahan2017communication, kairouz2019advances, bonawitz2019towards} is a distributed machine learning paradigm where training data resides
at autonomous client machines and the learning process is facilitated by a central server. The server maintains a shared model and alternates between requesting clients to try and improve it and integrating their suggested improvements back into that shared model.

A few interesting challenges arise from this model. First, the need for communication efficiency, both in terms of the size of data transferred and the number of required messages for reaching convergence. Second, clients are outside of the control of the server and as such may be unreliable, or even malicious. Third, while classical learning models generally assume that data is homogeneous, here privacy and the aforementioned communication concerns force us to deal with the data as it is seen by the clients; that is 1) \emph{non-IID} (identically and independently distributed) -- data may depend on the client it resides at, and 2)  \emph{unbalanced} -- different clients may possess different amounts of data.

In previous works \citep{DBLP:journals/corr/abs-1906-06629, NIPS2018_7712, li2019rsa, haddadpour2019convergence, pillutla2019robust}, unbalancedness is either ignored or is represented by a collection of a priori known \emph{client importance weights} that are usually derived from the amount of data each client has. This work investigates aspects that stem from this unbalancedness. Concretely, we focus on the case where unreliable clients declare the amount of data they have and may thus adversely influence their importance weight. We show that without some mitigation, a single malicious client can obstruct convergence in this manner even in the presence of popular FL defense mechanisms. Our experiments consider protections that replace the server step by a robust mean estimator, such as median \citep{10.1145/3154503, pmlr-v80-yin18a, chen2019distributed} and trimmed mean \citep{pmlr-v80-yin18a}.

The rest of this paper is organized as follows. In Section \ref{sec:setup}, we present required definitions and formalize the problem  addressed by this work. Section \ref{sec:preprocessing} presents our truncation-based preprocessing method and proves that it can be applied to a randomly-selected sample of client weights . In Section \ref{sec:evaluation}, we report on the results of our empirical evaluation. Conclusions and directions for future work are presented in Section \ref{sec:conclusion}.

\section{Problem setup}\label{sec:setup}

\subsection{Optimization goal} 

We are given $K$ clients where each client $k$ has a local collection $Z_k$ of $n_k$ samples taken IID\ from some unknown distribution over sample space $\bm{Z}$. We denote the unified sample collection as $Z = \bigcup_{k \in [K]} Z_k$ and the total number of samples as $n$ (i.e., $n = |Z| = \sum_{k \in [K]} n_k$) . 
Our objective is \emph{global} empirical risk minimization (ERM) for some loss function class $ \ell(w; \cdot)\colon \bm{Z} \to \mathbb{R}$, parameterized by $w \in \mathbb{R}^d$ \footnote{We note that some previous FL works specify a more generic finite-sum objective \citep{mcmahan2017communication}. However, this work investigates client-declared sample sizes, whose meaning is clear under the ERM interpretation but seems meaningless in  the finite-sum objective setting.}:

\begin{gather} \label{eq:goal}
 \min_{w \in \mathbb{R}^d} F(w), \text{ where } F(w) \coloneqq \frac{1}{n} \sum_{z \in Z} \ell(w ; z).
\end{gather}

In the following sections we denote the vector of client sample sizes as $\boldsymbol{N}= (n_1, n_2, \dots, n_K)$ and assume, w.l.o.g., that it is sorted in increasing order.

\subsection{Collaboration model} \label{subs:collab}

We restrict ourselves to the FL paradigm, which leaves the training data distributed on client machines, and learns a shared model by iterating between client updates and server aggregation.

Additionally, a subset of the clients, marked $\mathcal{B}$, can be Byzantine, meaning they can send arbitrary and possibly malicious results on their local updates. Moreover, unlike previous works, we also consider clients' sample sizes to be unreliable because they are reported by possibly Byzantine clients. When the distinction is important, values that are sent by clients are marked with an overdot to signify that they are unreliable (e.g.,  $\dot{n}_k$), while values that have been preprocessed in some way are marked with a tilde (e.g., $\tilde{n}_k$).

\subsection{Federated learning meta algorithm}

We build upon the baseline federated averaging algorithm ($FedAvg$) described by \citet{mcmahan2017communication}. There, it is suggested that in order to save communication rounds clients perform multiple stochastic gradient descent (SGD) steps while a central server occasionally averages the parameter vectors.

The intuition behind this approach becomes clearer when we mark the $k$\nth client's ERM objective function by $F_k(w) \coloneqq \frac{1}{n_k} \sum_{z \in Z_k} \ell(w ; z)$ and observe that the objective function in equation \eqref{eq:goal} can be rewritten as a weighted average of clients' objectives:

\begin{equation} \label{eq:goal-weighted}
F(w) \coloneqq \frac{1}{n} \sum_{k \in [K]} n_k F_k(w).
\end{equation}

Similarly to previous works \citep{pillutla2019robust, chen2019communicationefficient, chen2019distributed}, we capture a large set of algorithms by abstracting $FedAvg$ into a meta-algorithm for FL (Algorithm \ref{algo:meta}). We require three procedures to be specified by any concrete algorithm: 

\begin{enumerate}[label=(\alph*)]
\item $Preprocess$ -- receives, possibly byzantine, $\dot{n}_k$'s from clients and produces secure estimates marked as $\tilde{n}_k$'s. To the best of our knowledge, previous works ignore this procedure and assume $n_k$'s are correct.
\item $ClientUpdate$ -- per-client $w_k$ computation. In $FedAvg$, this corresponds to a few local mini-batch SGD rounds. See Algorithm \ref{algo:fedavg-clientupdate} for pseudocode.

\item $Aggregate$ -- the server's strategy for updating $w$. In $FedAvg$, this corresponds to the weighted arithmetic mean, i.e., $w \gets \frac{1}{\dot{n}} \sum_{k \in [K]} \dot{n}_k \dot{w}_k$.
\end{enumerate}

\begin{minipage}{0.45\linewidth}
\vskip -0.9in
\begin{algorithm}[H]
\centering
\caption{$FedAvg$: $ClientUpdate$} 
\label{algo:fedavg-clientupdate} 
\begin{flushleft}
        \textbf{Hyperparameters:} learning rate ($\eta$), number of epochs ($E$), and batch size ($B$).
\end{flushleft}
\begin{algorithmic}[1]
\FOR{$E$ epochs}
    \FOR{$B$-sized batch $\mathcal{B} \subseteq Z_k$}
        \STATE $w_k \leftarrow w_k - \eta \frac{1}{B} \sum_{z \in \mathcal{B}}  \nabla \ell (w_k; z)$
    \ENDFOR
\ENDFOR
\end{algorithmic}
\end{algorithm}
\end{minipage}
\hfill
\begin{minipage}{0.54\linewidth}
\begin{algorithm}[H]
\centering
\caption{Federated Learning Meta-Algorithm} 
\label{algo:meta} 
\begin{flushleft}
        \textbf{Given procedures:} $Preprocess$, $ClientUpdate$, and $Aggregate$.
\end{flushleft}
\begin{algorithmic}[1]
    \STATE $\{\dot{n}_k\}_{k \in [K]} \gets$ collect sample size from each client
    \STATE $\{\tilde{n}_k\}_{k \in [K]} \gets Preprocess(\{\dot{n}_k\}_{k \in [K]})$
    \STATE $w \gets $ initial guess
    \FOR{$t \gets 1$ \textbf{to} $T$}
        \STATE $S_t \gets$ a random set of client indices
        \FORALL{$k \in S_t$}
            \STATE $\dot{w}_k \gets ClientUpdate(\tilde{n}_k, w)$ 
        \ENDFOR
        \STATE $w \gets Aggregate(\{\langle \tilde{n}_k, \dot{w}_k \rangle \}_{k \in S_t})$
    \ENDFOR
\end{algorithmic}
\end{algorithm}
\end{minipage}

\section{Preprocessing client-declared sample sizes}\label{sec:preprocessing}

\subsection{Preliminaries} 

The following assumption is common among works on Byzantine robustness: 

\begin{asm}[Bounded Byzantine proportion]\label{asm:1}
 The proportion of clients who are Byzantine is bounded by some constant $\alpha$; i.e., $\frac{1}{K}|\mathcal{B}| \le \alpha$.
\end{asm}

The next assumption is a natural generalization when considering unbalancedness: 

\begin{asm}[Bounded Byzantine weight proportion]\label{asm:2}
The proportion between the combined weight of Byzantine clients and the total weight is bounded by some constant $\alpha^*$; i.e., $\frac{1}{n}\sum_{b \in \mathcal{B}} n_b  \le \alpha^*$.
\end{asm}

Previous works on robust aggregation \citep{DBLP:journals/corr/abs-1906-06629, NIPS2018_7712, li2019rsa, haddadpour2019convergence, zhao2018federated} either used Assumption \ref{asm:1}, without considering the unbalancedness, or implicitly used Assumption \ref{asm:2}. However, we observe that Assumption \ref{asm:2} is unattainable in practice since Byzantine clients can often influence their weight thus rendering their weight proportion unbounded.
% Also note that while it is tempting to directly use the weighted values with a robust aggregation (i.e.,  $Aggregate(\{n_k * w_k \}_{k \in S_t})$), it generally does not produce a useful mean estimation (it estimates the weighed sum divided by the number of addends).

We address this gap with the following definition and an appropriate $Preprocess$ procedure.

\begin{definition}[$\mwp$] 
Given a proportion $p$, and a vector $\bm{V}=(v_1,...,v_{|\bm{V}|})$ sorted in increasing order, the \emph{maximal weight proportion}, $\mwp(\bm{V}, p)$, is the maximum combined weight for \emph{any} $p$-proportion of the values of $\bm{V}$:

% \YT{The origin of $v_i$ was unclear, so I specified that $\bm{V}$ was a vector of $v_i$'s.}
% OK thanks

% By defining $\topa(\bm{N}, p)$ as the collection of the $p K$ largest values in $\bm{N}$ we can formulate this as:

$$
\mwp(\bm{V}, p) \coloneqq \frac{1}{
\sum_{v \in \bm{V}} v
}\sum_{(1 - p)|\bm{V}| < i} v_i.
$$

% \YT{Isn't $|\bm{V}|$ supposes to be $\bm{\sum} \bm{V}$?}  yes thanks

Note that this is just the weight proportion of the $p |\bm{V}|$ clients with the largest sample sizes.

%  \YT{Replaced $K$ with $|\bm{V}|$.} thanks

\end{definition}

In the rest of this work will assume Assumption \ref{asm:1} and design a $Preprocess$ procedure that ensures the following:

\begin{equation} \label{eq:preproc} 
\mwp(Preprocess(\bm{N}), \alpha) \leq \alpha^*.
\end{equation}

Observe that this requirement enables the use of weighted robust mean estimators in a realistic setting by ensuring that Assumption \ref{asm:2} holds for the preprocessed client sample sizes. Also note that here, $\alpha$ is our assumption about the proportion of Byzantine clients while $\alpha^*$ relates to an analytical property of the underlying robust algorithm. For example, we may replace the federated average with a weighted median as suggested by \citet{10.1145/3154503}, in which case, $\alpha^*$ must be less than $1/2$.

\subsection{Truncating the values of N}

Our suggested preprocessing procedure uses element-wise truncation of the values of $\bm{N}$ by some value $U$, marked $\trunc(\bm{N}, U) = (\text{min}(n_1, U), \text{min}(n_2, U), \dots, \text{min}(n_K, U))$. Given $\alpha$ and $\alpha^*$, we search for the maximal truncation which satisfies \eqref{eq:preproc}:

\begin{equation} \label{eq:trunc} 
    U^* \coloneqq \argmax_{U \in \mathbb{N}}\ s.t.\ \mwp(\trunc(\bm{N}, U), \alpha) \leq \alpha^*
\end{equation}

%  \\
%     \mwp(\trunc(\bm{N}, U), \alpha) \leq \alpha^*.

Here $\alpha$ and $U^*$ present a trade-off. Higher $\alpha$ means more Byzantine tolerance but requires smaller truncation value $U^*$, which, may cause slower and less accurate convergence, as we demonstrate empirically and theoretically in Theorem \ref{thm:dist-bound}.

We note that given $\alpha$ and $\alpha^*$, truncating $\bm{N}$ by the solution of \eqref{eq:trunc} is optimal in the sense that any other $Preprocess$ procedure that adheres to \eqref{eq:preproc} has an equal or larger $\bm{L}^1$ distance from the original $\bm{N}$. This follows immediately from the observation that, when truncating the values of $\bm{N}$, the entire distance is due to the truncated elements and if there was another applicable vector closer to $\bm{N}$, we could have redistributed the difference to the largest elements and increase $U^*$ in contradiction to its maximally. 

% \YT{I remember Amit explaining this to me, and me understanding this hypothesis, but I can't seem to wrap my head around it now.}
% added a few words

\subsubsection{Finding U given alpha}

If one has an estimate for $\alpha$ it is easy to calculate $U^*$. For example, by going over values in $\bm{N}$ in a decreasing order (i.e., from index $K$ downwards) until finding a value that satisfies the inequality in \eqref{eq:trunc}. Then we mark the index of this value by $u$ and use the fact that in the range $[n_u, n_{u+1}]$ we can express $\mwp(\trunc(\bm{N}, U), \alpha)$ as a simple function of the form $\frac{a + b U}{c + d U}$:

$$
\frac{\sum_{(1-\alpha)K < i \leq u} n_i + |\{n_i : i > \max(u, (1-\alpha)K) \}| U}{\sum_{i \leq u} n_i + |\{n_i : i > u\}| U},
$$

for which we can solve \eqref{eq:trunc} with

\begin{equation} \label{eq:3}
U^* \gets \left\lfloor \frac{a - c \alpha^*}{d \alpha^* -b} \right\rfloor.
\end{equation}

\subsubsection{The alpha-U trade-off}

When we do not know $\alpha$, as a practical procedure, we suggest plotting $U^*$ as a function of $\alpha$. In order to do so we can start with $\alpha \gets \alpha^*$, $U \gets n_1$, and alternate between decreasing $\alpha$ by $1/K$ (one less Byzantine client tolerated) and solving \eqref{eq:trunc}. This procedure can be made efficient by saving intermediate sums and using a specialized data structure for trimmed collections.  See Algorithm \ref{algo:report} for pseudocode and Figure \ref{fig:tradeoff} for an example output.

\begin{minipage}{0.49\linewidth}
\captionsetup{type=figure}
\includegraphics[width=\linewidth]{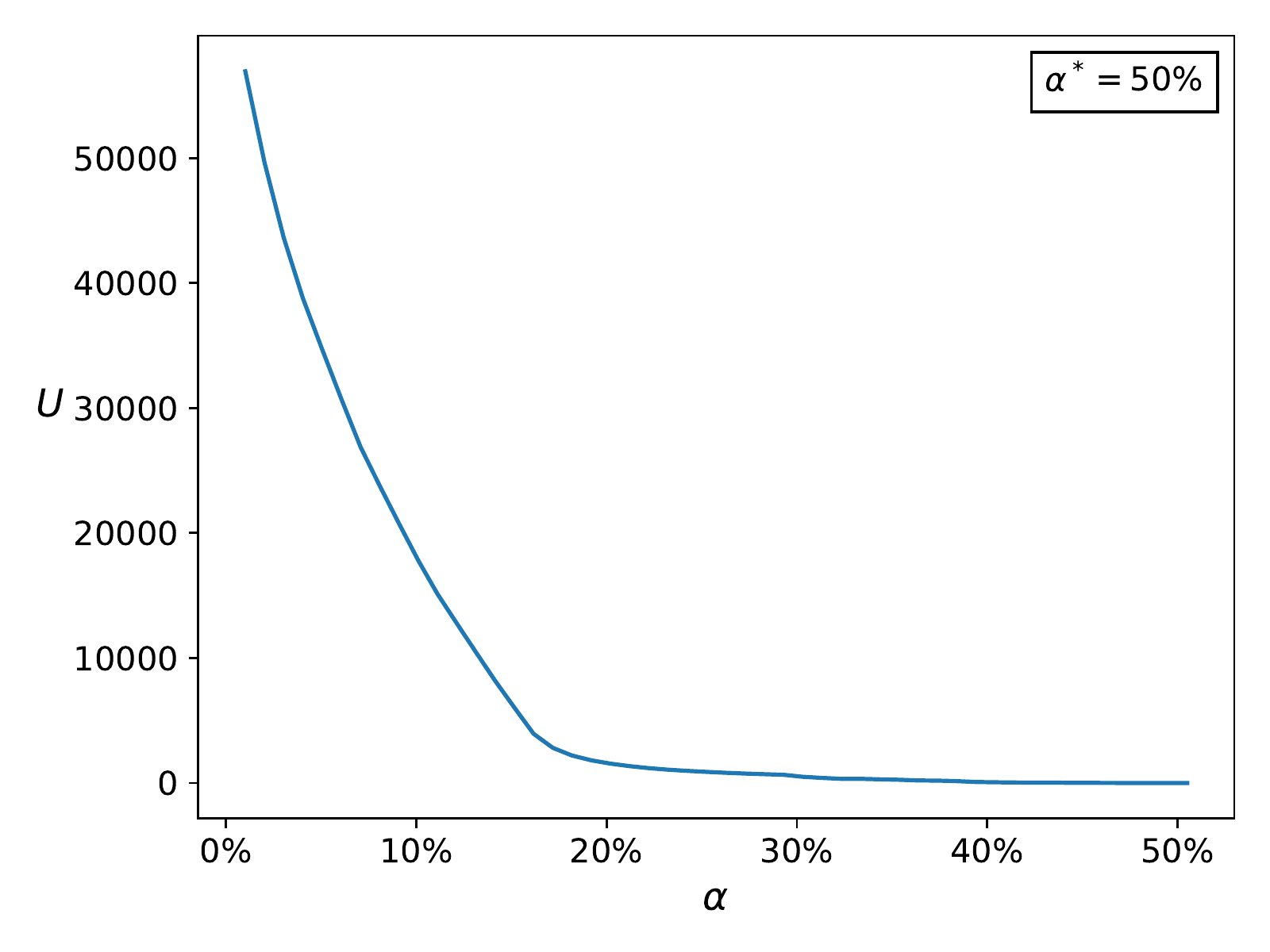}
\captionof{figure}{Example plot of data generated by executing Algorithm \ref{algo:report} on unbalanced vector $\bm{N}$ and $\alpha^* = 50\%$ (this vector corresponds to the partition used in our experiments; See Section \ref{Experimental_Setup} for details).}
\label{fig:tradeoff}
\end{minipage}
\hfill
\begin{minipage}{0.49\linewidth}
\vskip -1.4in
\begin{algorithm}[H]
\centering
\caption{Report ($\alpha$, $U^*$) Pairs} 
\label{algo:report} 
\begin{algorithmic}
\STATE $\alpha \gets \alpha^*$ 
\FOR{$u \gets 1$ \textbf{to} $K-1$}
    \WHILE{$\mwp(\trunc(\bm{N}, n_{u+1}), \alpha) > \alpha^*$}
        \STATE $U^* \gets $ solve \eqref{eq:3} for $U \in [n_u, n_{u+1}]$
        \STATE \textbf{report} ($\alpha$, $U^*$)        
        \STATE $\alpha \gets \alpha - \frac{1}{K}$
    \ENDWHILE
\ENDFOR
\end{algorithmic}
\end{algorithm}
\end{minipage}

\subsection{Truncation given a partial view of N}

When K is very large we may want to sample only $k \ll K$ elements IID from $\bm{N}$. 
In this case we will need to test that the inequality in \eqref{eq:trunc} holds with high probability.

We consider $k$ discrete random variables taken IID from $\bm{N}$ after truncation by $U$, that is, taken from a distribution over $\{0, 1, \dots, U\}$. We mark these random variables as $X_1, X_2, \dots, X_k$, and their order statistic as $X_{(1)}, X_{(2)}, \dots, X_{(k)}$ where $X_{(1)} \le X_{(2)} \le \dots \le X_{(k)}$.

\begin{theorem}\label{thm:partial-view}
Given parameter $\delta > 0$ and $\varepsilon_1 = \sqrt{\frac{\ln{(3/\delta)}}{2 k}}$, $\varepsilon_2 = U \sqrt{\frac{\ln{\ln{(3/\delta)}}}{2 (k(\alpha - \varepsilon_1) + 1)}}$, $\varepsilon_3 = U \sqrt{\frac{\ln{\ln{(3/\delta)}}}{2 k}}$, we have that $\mwp(\trunc(\bm{N}, U), \alpha) \le \alpha^*$ is true with $1 - \delta$ confidence if the following holds:

\begin{equation}
\begin{split}
\frac{\alpha \big(\frac{\sum^k_{i \gets \lceil(1 - (\alpha  - \varepsilon_1))k\rceil} X_{(i)}}{k - \lceil(1 - (\alpha  - \varepsilon_1))k\rceil + 1} + \varepsilon_2 \big) }{\big(\frac{1}{k} \sum_{i \in [k]} X_i - \varepsilon_3\big)}
 \le \alpha^*
\end{split}
\end{equation}

%  $ k \geq \max \displaystyle (\frac{\ln{\frac{3}{\delta}}}{2\varepsilon_1^2}, \frac{\frac{U^2 \ln{\frac{3}{\delta}}}{ 2\varepsilon_2^2} - 1}{\alpha - \varepsilon_1}, \frac{U^2 \ln{\frac{3}{\delta}}}{2\varepsilon_3^2})$

\end{theorem}

\begin{proof}
See Appendix \ref{prf:partial-view}
\end{proof}

\subsection{Convergence analysis}

After applying our $Preprocess$ procedure we have the truncated number of samples per client, marked $\{\tilde{n}_k\}_{k \in [K]}$. We can trivially ensure that any algorithm instance works as expected by requiring that clients ignore samples that were truncated. That is, even if an honest client $k$ (non-Byzantine) has $n_k$ samples it may use only $\tilde{n}_k$ samples during its $ClienUpdate$.

Although this solution always preserves the semantics of 
\emph{any} underlying algorithm, it does hurt convergence guarantees since the total number of samples decreases [\citealt[Tables 5 and 6]{kairouz2019advances}; \citealt{pmlr-v80-yin18a}; \citealt{haddadpour2019convergence}]. Interestingly, \citet[Theorem~3]{li2019convergence} analyze the baseline \emph{FedAvg} and show that its convergence bound decreases with $max\,n_k$ (marked there as $\nu$). This suggests that in some cases truncation itself may mitigate the decrease in total sample size.

Additionally, we note that in practice, the performance of federated averaging based algorithms improves when honest clients use all their original $n_k$ samples. Intuitively, this follows easily from the observation that $Aggregate$ procedures are generally composite mean estimators and $ClientUpdate$ calls are likely to produce more accurate results given more samples.

Lastly, as we have mentioned before, convergence is guaranteed, but we note that the optimization goal itself is inevitably skewed in our Byzantine scenario. The following theorem bounds this difference between original weighted optimization goal \eqref{eq:goal-weighted} and the new goal after truncation.
In order to emphasize the necessity of this bound (in terms of Assumption \ref{asm:2}), we use overdot and tilde to signify unreliable and truncated values, respectively, as previously described in Subsection \ref{subs:collab}.

\begin{theorem}\label{thm:dist-bound}
Given the same setup as in \eqref{eq:goal} and a truncation bound $U$, the following holds for all $w \in \mathbb{R}^d$:
\begin{equation*}
\begin{split}
    &
    \norm{
    \frac{1}{\dot{n}}\sum_{i \in [K]} \dot{n}_i F_i(w) -
    \frac{1}{\tilde{n}}\sum_{i \in [K]} \tilde{n}_i F_i(w)} \le \\
    &
    \norm{
    \sum_{i : \dot{n}_i > U} \big(\frac{\dot{n}_i}{\dot{n}} - \frac{1}{K}\big) F_i(w) +
    \big(\frac{1}{\dot{n}} - \frac{1}{\tilde{n}} \big) \sum_{i : \dot{n}_i \le U}  \mathcal{L}(Z_i)}
\end{split}
\end{equation*}

% $\dot{p}_i$ and $\tilde{p}_i$ are the relative proportion from the samples that client $i$ in the unreliable and truncated case respectively (i.e., ,  
Where $\mathcal{L}(Z_i)$ is defined as $\sum_{z \in Z_i} \ell(w ; z)$.
\end{theorem}

\begin{proof}
See Appendix \ref{prf:dist-bound}
\end{proof}

From the bound in Theorem \ref{thm:dist-bound} we can clearly see how the coefficients in the left term, $(\dot{n}_i / \dot{n} - 1 / K)$, stem from unbalancedness in the values above the truncation threshold while the coefficient in the right term, $(1 / \dot{n} - 1 / \tilde{n})$, accounts for the increase of relative weight of the values below the truncation threshold. Additionally, note that this formulation demonstrates how a single Byzantine client can increase this difference arbitrarily by increasing its $\dot{n}_i$. Lastly, observe how both terms vanish as $U$ increases, which motivates our selection of $U^*$ as the \emph{maximal} truncation threshold for any given $\alpha$ and $\alpha^*$.

% for any definition of $ClienUpdate$, it uses fewer samples than available thus hurt convergence rate. 

% Our experiments show that honest client $k$ can in fact use all its $n_k$ samples during $ClientUpdate$ and this slightly improves the convergence rate.

% elementry way to asert option that gerentee that $ClienUpdate$ and $Aggregate$

\section{Evaluation}\label{sec:evaluation}

In this section we demonstrate how truncating $\bm{N}$ is a crucial requirement for Byzantine robustness. That is, we show that no matter what is the specific attack or aggregation method, using $\bm{N}$ ``as-is" categorically devoids any robustness guaranties.

\label{p:code}

The code for the experiments is based on the Tensorflow machine learning library \citep{tensorflow2015-whitepaper}. Specifically, the code for the shakespeare experiments is based on the Tensorflow Federated sub-library of Tensorflow. It is given under the Apache license 2.0. Our code can be found in \url{https://github.com/amitport/Towards-Federated-Learning-With-Byzantine-Robust-Client-Weighting}. We perform the experiments using a single NVIDIA GeForce RTX 2080 Ti GPU, but the results are easily reproducible on any device.

\subsection{Experimental setup} \label{Experimental_Setup}

\subsubsection{The machine learning tasks and models}

% We run our experiments on two relatively simple machine learning tasks: MNIST with a synthetic partition and Shakespeare. The first is the standard MNIST  image classification task \citep{lecun2010mnist} with a synthetic distribution of the samples among the clients.  The second task is taken from the LEAF benchmark for Federated Settings \citep{caldas2019leaf} and represents a more realistic unbalanced distribution of the samples among the clients.

\begin{figure*}
\centering
\includegraphics[width=\linewidth]{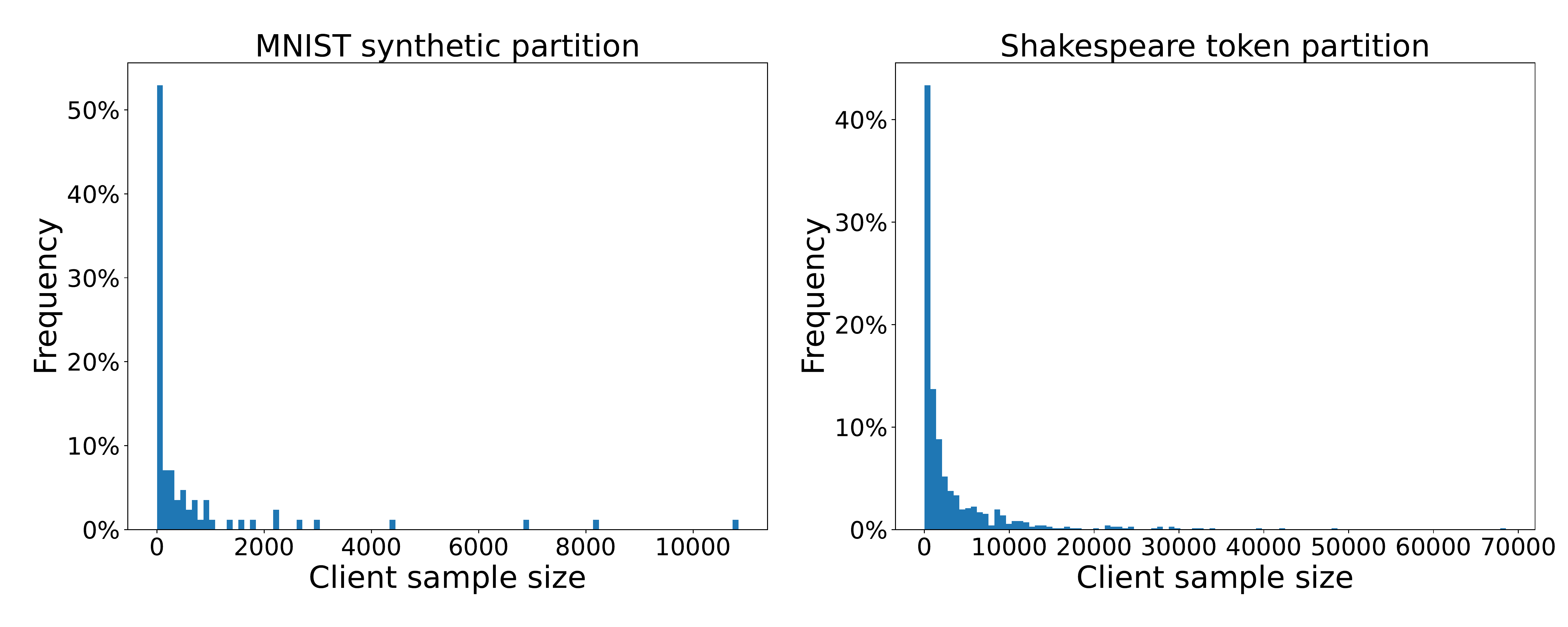}
\caption{Histogram of the sample partitions of the MNIST (left) and Shakespeare (right) datasets.}
\label{fig:partition}
\end{figure*}

\paragraph{Shakespeare: next-character-prediction partitioned by speaker.}

Presented in the original \textit{FedAvg} paper \citep{mcmahan2017communication} and also part of the LEAF benchmark \citep{caldas2019leaf}, the Shakespeare dataset contains 422,615 sentences taken from \emph{The Complete Works of William Shakespeare} \citep{shakespeare1996complete} (freely available public domain texts). The next-character-prediction task with the per-speaker partitioning represents a realistic scenario in the FL domain. Each client trains using an LSTM recurrent model  \citep{Hochreiter1997LongSM} with hyperparameters matching those suggested by \citet{reddi2020adaptive} for \emph{FedAvg}.

% For our first task \YT{We say 'for our first task' for MNIST, bue there is no such wording for Shakespeare}
% thanks

\paragraph{MNIST: digit recognition with synthetic client partitioning.}
 The MNIST database \citep{lecun2010mnist} (available under Creative Commons Attribution-ShareAlike 3.0 license) includes $28{\times}28$ grayscale labeled images of handwritten digits split into a 60,000 training set and a 10,000 test set. We randomly partition the training set among 100 clients. The partition sizes are determined by taking 100 samples from a Lognormal distribution with $\mu\!=\!1.5$, $\sigma\!=\!3.45$, and then interpolating corresponding integers that sum to 60,000. This produces a right-skewed, fat-tailed partition size distribution that emphasizes the importance of correctly weighting aggregation rules and the effects of truncation. Clients train a classifier using a 64-units perceptron with RelU activation and $20\%$ dropout, followed by a softmax layer. Following \citet{pmlr-v80-yin18a}, on every communication round, all clients perform mini-batch SGD with 10\% of their examples. 

Note that the Shakespeare and MNIST synthetic tasks were selected because they are relatively simple, unbalanced tasks. Simple, because we want to evaluate 
a preprocessing phase and avoid tuning of the underlying algorithms we compare. Unbalanced, since as can be understood from Theorem \ref{thm:dist-bound}, when the client sample sizes are spread mostly evenly, ignoring the client sample size altogether is a viable approach. See Figure \ref{fig:partition} for the histograms of the partitions.

\subsubsection{The server}

We show three $Aggregate$ procedures. Arithmetic mean, as used by the original $FedAvg$, and two additional procedures that replace the arithmetic mean with robust mean estimators. The first of the latter uses the coordinatewise median \citep{10.1145/3154503, pmlr-v80-yin18a}. That is, each server model coordinate is taken as the median of the clients' corresponding coordinates. The second robust aggregation method uses the coordinatewise trimmed mean \citep{pmlr-v80-yin18a} that, for a given hyperparameter $\beta$, first removes $\beta$-proportion lowest and $\beta$-proportion highest values in each coordinate and only then calculates the arithmetic mean of the remaining values.

% These are described in \citep{10.1145/3154503} and \citet{pmlr-v80-yin18a} respectively.

% These robust aggregation methods just replace the server aggregation step with mean estimators that more resilient to outlier values.

When preprocessing the client-declared sample size, we compare three options: We either ignore client sample size, truncate according to $\alpha\!=\!10\%$ and $\alpha^*\!=\!50\%$, or just passthrough client sample size as reported.

\subsubsection{The clients and attackers}

For the Shakespeare experiments, we examine a \emph{model negation attack} \citep{NIPS2017_6617}. In this attack, each attacker ``pushes" the model towards zero by always returning a negation of the server's model. When the data distribution is balanced, this attack is easily neutralized since Byzantine clients typically send extreme values. However, in our unbalanced case, we demonstrate that without our preprocessing step, this attack cannot be mitigated even by robust aggregation methods.

For MNIST, in order to provide comparability, we follow the experiment shown by \citet{pmlr-v80-yin18a} in which $10\%$ of the clients use a \emph{label shifting attack}. In this attack, Byzantine clients train normally except for the fact that they replace every training label $y$ with $9\!-\!y$. The values sent by these clients are then incorrect but are relatively moderate in value making their attack somewhat harder to detect. This is in addition to the model negation attacks, already shown in the Shakespeare experiments.

We first execute our experiment without any attacks for every server aggregation and preprocessing combination. Then, for each attack type, we repeat the process two additional times: 1) with a single attacker that declares 10 million samples, and 2) with $10\%$ attackers that declare 1 million samples each.

% \AP{Are the numbers still accurate?} yes

% three times for every server aggregation, preprocessing, and attack combination: (1) without attackers, (2) with a single attacker that declares 10 million samples, or (3) with $10\%$ attackers that declare 1 million samples each.

% We use a very simple, easily neutralized attack that demonstrate the imperativeness of preprocessing. On every round, each attacker ``pushes" the model towards zero by always returning a negation of the server's model. 

% \begin{figure*}
% % \vskip 0.2in
% % \begin{center}
% % ,width=14cm,height=12cm
% \centerline{\includegraphics[width=\linewidth]{model_negation_attack}}
% \caption{Accuracy by round under model negation attack.
% }
% \label{fig:model_negation_attack}
% % \end{center}
% % \vskip -0.2in
% \end{figure*}

\subsection{Results}

\begin{figure*}[!t]
\centering
\includegraphics[width=\linewidth]{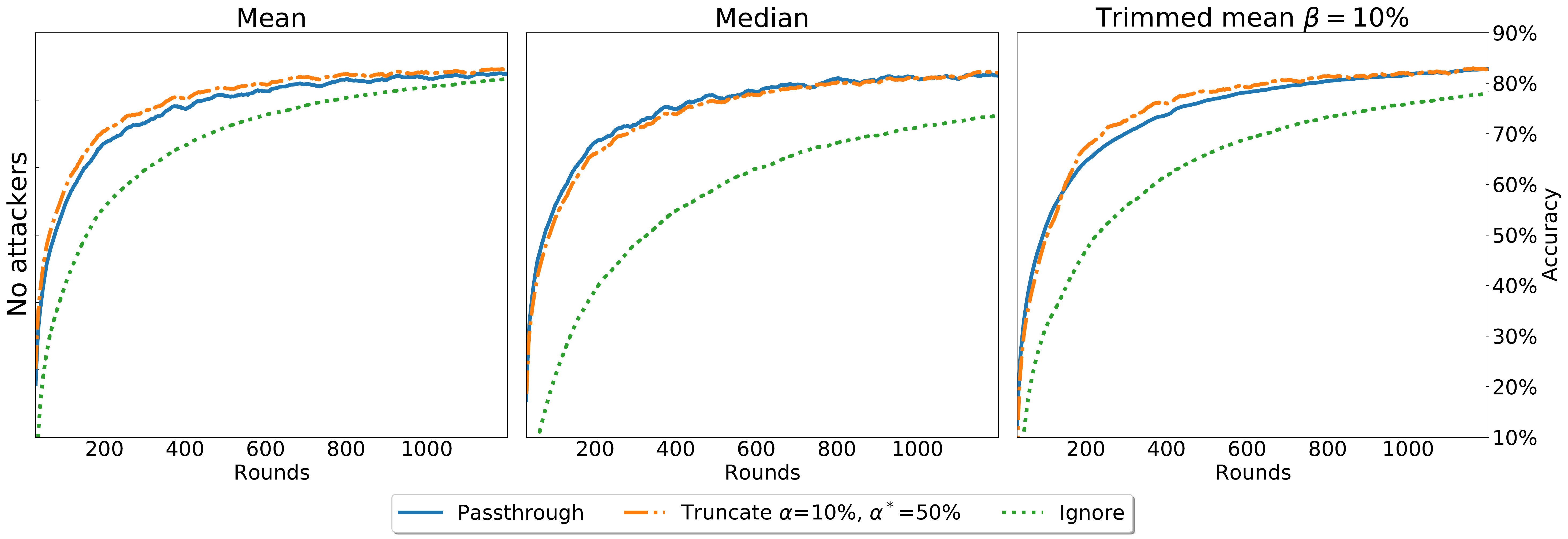}
\caption{Accuracy by round without any attackers for the Shakespeare experiments. Curves correspond to preprocessing procedures and columns correspond to different aggregation methods. It can be seen that our method (dashed orange curve) remains comparable to the properly weighted mean estimators (solid blue curve) while ignoring clients' sample sizes (dotted green curve) is sub-optimal. This effect is pronounced when the unweighted median is used, since with our unbalanced partition it is generally very far from the mean. Figure \ref{fig:mnist:no_attackers} in Appendix \ref{sec:mnist} shows similar results for the MNIST experiments.}
\label{fig:no_attackers}
\end{figure*}

\begin{figure*}
\centering
\includegraphics[width=\linewidth]{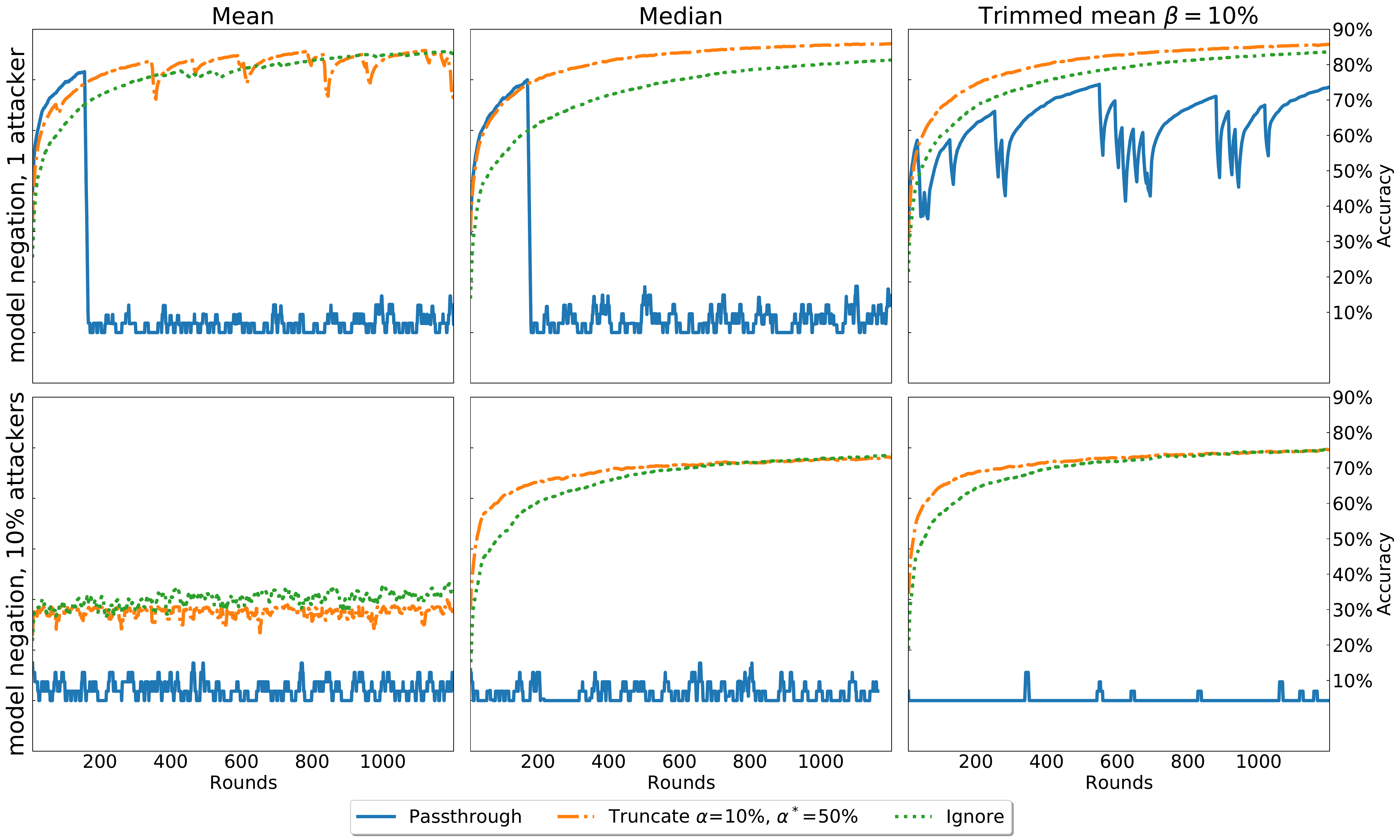}
\caption{Accuracy by round under Byzantine attacks for the Shakespeare experiments. Curves correspond to preprocessing procedures and columns correspond to different aggregation methods. In the two rows of the experiment the Byzantine clients perform a model negation attack with one and $10\%$ attackers, respectively.
\\\hspace{\linewidth}
We observe that even with a single attacker performing a trivial attack (first row), using the weights directly (solid blue curve) is devastating while when our preprocessing method is used in conjunction with robust mean aggregations (dashed orange curve, two last columns) convergence remains stable even when there are actual $\alpha$ (=$10\%$) attackers (second row). We note that in some cases our method may be slightly less efficient compared with the preprocessing method that ignores sample size altogether (dotted green curve, second row, leftmost column). This is to be expected because we allow Byzantine clients to potentially get close to $\alpha^*$-proportion ($50\%$, in this case) of the weight. However, our method is significantly closer to the optimal solution when there are no or only a few attackers (see Figure \ref{fig:no_attackers}). Moreover, when used in conjunction with robust mean aggregation methods it maintains their robustness properties. Figure \ref{fig:mnist:attack} in Appendix \ref{sec:mnist} shows similar results for the MNIST experiments.}
\label{fig:attack}
\end{figure*}

The Shakespeare experiments without any attackers is shown in Figure \ref{fig:no_attackers} and the executions with attackers are shown in Figure \ref{fig:attack}. The results of the MNIST experiments were almost identical and are deferred to Appendix \ref{sec:mnist} for brevity.

The results from the first experiment, running without any attackers (Figure \ref{fig:no_attackers}), demonstrate that ignoring client sample size results in reduced accuracy, especially when median aggregation is used, whereas truncating according to our procedure is significantly better and is on par with properly using all weights. These results highlight the imperativeness of using sample size weights when performing server aggregations.

While Figure \ref{fig:no_attackers} shows that the truncation-based preprocessing performs on par with that of taking all weights into consideration when all clients are honest, Figure \ref{fig:attack} demonstrates that the results are very different when there is an attack. In this case, we see that when even a single attacker reports a highly exaggerated sample size and the server relies on all the values of $\bm{N}$, the performance of all aggregation methods including robust median and trimmed mean quickly degrade.

In contrast, in our experiments robustness is maintained when truncation-based preprocessing is used in conjunction with robust mean aggregations, even when  Byzantine clients attain the maximal supported proportion ($\alpha\!=\!10\%$).

\section{Conclusion and future work}\label{sec:conclusion}

Our method is based on truncating the weight values reported by clients in a manner that bounds from above the proportion $\alpha^*$ of weights that can be attributed to Byzantine clients, given an upper bound on the proportion of clients $\alpha$ that may be Byzantine. Different values of parameter $\alpha$ represent different points in the trade-off between model quality and Byzantine-robustness, where higher values increase robustness when attacks do occur but decrease convergence rate even in the lack of attacks.

We evaluated the performance of our truncation method empirically when applied as a preprocessing stage, prior to several aggregation methods. The results of our experiments establish that: 1) in the absence of attacks, model convergence is on par with that of properly using all reported weights, and 2) when attacks do occur, the performance of combining truncation-based preprocessing and robust aggregations incurs almost no penalty in comparison with the performance of using of all weights in the lack of attacks, whereas without preprocessing, even robust aggregation methods collapse to performance worse than that of a random classifier. 

When the number of clients is very large, performing server preprocessing and aggregation on the server may become computationally infeasible. We prove that, in this case, truncation-based preprocessing can achieve the same upper bound on $\alpha^*$ w.h.p. based on the weight values reported from a sufficiently large number of the clients selected IID.

As with many Byzantine-robust algorithms, the selection of $\alpha$ has a significant impact on the underlying model and, specifically, on fairness towards clients that hold underrepresented data, which may inadvertently be considered outliers. In future work, we plan to analyze further the trade-off between robustness and the usage of client sample size in rectifying data unbalancedness. We also plan to investigate alternative forms of estimating client importance that may avoid client sample size altogether.

% \YT{I'm not sure if this could still be considered a discussion on the negative societal impacts of the paper...}
% it's good enough in my opinion :) our work is not different from any other robust learning work, and doesn't need more than this 
\clearpage

% \begin{ack}

% \end{ack}

\bibliography{main}
\bibliographystyle{plainnat}

\appendix

\section*{Appendix}

\section{Proofs}\label{proofs}

\subsection{Proof of theorem \ref{thm:partial-view}}\label{prf:partial-view}

First, in the scope of this proof we use a couple of additional notations:
\begin{itemize}
\item $\topa(\bm{V}, p)$: The collection of $p |\bm{V}|$ largest values in $\bm{V}$.
\item  $\bm{\sum} \bm{V}$: The sum of all elements in $\bm{V}$.
\end{itemize}

We observe that $\mwp(\trunc(\bm{N}, U), \alpha) \le \alpha^*$ can be rewritten as
\begin{equation}
\begin{split}
&
\mwp(\trunc(\bm{N}, U), \alpha) = \frac{ \bm{\sum} \topa(\trunc(\bm{N}, U), \alpha) }{\bm{\sum} \trunc(\bm{N}, U)}
\label{eq:5}
= \frac{ \alpha \mathbb{E} [ \topa(\trunc(\bm{N}, U), \alpha) ] }{\mathbb{E} [\trunc(\bm{N}, U)]} \le \alpha^*
\end{split}
\end{equation}

Then we note that membership in $\topa(\trunc(\bm{N}, U), \alpha)$ can be viewed as a simple Bernoulli random variable with probability $\alpha$, for which we obtain the following bound using Hoeffding's inequality, $t\geq 0$:

\begin{equation}
\begin{split}
\Pr \big[|\{i \in [k] : X_i \in \topa(\trunc(\bm{N}, U), \alpha)\}| \leq (\alpha - t)k\big]
\leq \mathrm{e}^{-2 t^2 k}
\end{split}
\end{equation}

Therefore with $t = \varepsilon_1$, we have the following with $1 - \frac{\delta}{3}$ confidence:

\begin{equation}
\begin{split}
\bm{\sum} \topa(\{X_i\ |\ i \in [k]\}, \alpha)
\label{eq:6}
\leq  
\bm{\sum} \{X_{(i)}\ |\ \lceil(1 - (\alpha  - \varepsilon_1))k\rceil \le i \le k\}
\end{split}
\end{equation}

% \sum^k_{i \gets \lceil(1 - (\alpha  - \varepsilon_1))k\rceil} X_i

Using Hoeffding's inequality again, we can bound the expectation of $X_{(i)}\ |\ \lceil(1 - (\alpha  - \varepsilon_1))k\rceil \le i \le k$
by $\varepsilon_2$ with $1 - \frac{\delta}{3}$ confidence and together with \eqref{eq:6} have that:

\begin{equation}
\begin{split}
\label{eq:7}
\textstyle \mathbb{E} [ \topa(\trunc(\bm{N}, U), \alpha) ]
\leq  \frac{\sum^k_{i \gets \lceil(1 - (\alpha  - \varepsilon_1))k\rceil} X_{(i)}}{k - \lceil(1 - (\alpha  - \varepsilon_1))k\rceil + 1} + \varepsilon_2
\end{split}
\end{equation}

Then, using Hoeffding's inequality for the third time, $\mathbb{E} [\trunc(\bm{N}, U)]$ is bound from below by $\varepsilon_3$ with $1 - \frac{\delta}{3}$ confidence:

\begin{equation}
\begin{split}
\label{eq:8}
\mathbb{E} [\trunc(\bm{N}, U)] \geq \frac{1}{k} \sum_{i \in [k]} X_i - \varepsilon_3
\end{split}
\end{equation}

The proof is concluded by applying (\ref{eq:6}-\ref{eq:8}) to \eqref{eq:5} using the union bound.

% we get 

% \begin{align*}
% \Pr \Bigg[\topa(\trunc(\bm{N}, U), \alpha) > \\
% \frac{\alpha \big(\frac{\sum^k_{i \gets \lceil(1 - (\alpha  - \varepsilon_1))k\rceil} X_i}{k - \lceil(1 - (\alpha  - \varepsilon_1))k\rceil + 1} + \varepsilon_2 \big)
% }{ \frac{1}{k} \sum_{i \in [k]} X_i - \varepsilon_3} \Bigg] \le \delta
% \end{align*}

\subsection{Proof of theorem \ref{thm:dist-bound}}\label{prf:dist-bound}

Using the fact that $\tilde{n} \leq U K$ we get:
\begin{equation*}
\begin{split}
& \norm{
    \frac{1}{\dot{n}}\sum_{i \in [K]} \dot{n}_i F_i(w) -
    \frac{1}{\tilde{n}}\sum_{i \in [K]} \tilde{n}_i F_i(w)} = \\ &
    \norm{
     \sum_{i : \dot{n}_i > U} \frac{\dot{n}_i}{\dot{n}} F_i(w) +
    \frac{1}{\dot{n}} \sum_{i : \dot{n}_i \le U} \mathcal{L}(Z_i)
    -  \sum_{i : \dot{n}_i > U} \frac{U}{\tilde{n}} F_i(w)
    - \frac{1}{\tilde{n}} \sum_{i : \dot{n}_i \le U} \mathcal{L}(Z_i)} \leq \\ &
        \norm{
     \sum_{i : \dot{n}_i > U} \frac{\dot{n}_i}{\dot{n}} F_i(w) +
    \frac{1}{\dot{n}} \sum_{i : \dot{n}_i \le U} \mathcal{L}(Z_i)
    - \sum_{i : \dot{n}_i > U} \frac{1}{K} F_i(w)
    - \frac{1}{\tilde{n}} \sum_{i : \dot{n}_i \le U} \mathcal{L}(Z_i)} = \\ &
    \norm{
    \sum_{i : \dot{n}_i > U} \big(\frac{\dot{n}_i}{\dot{n}} - \frac{1}{K}\big) F_i(w) +
    \big(\frac{1}{\dot{n}} - \frac{1}{\tilde{n}} \big) \sum_{i : \dot{n}_i \le U}  \mathcal{L}(Z_i)}
\end{split}
\end{equation*}

\clearpage

\section{MNIST experiment results}\label{sec:mnist}

This section provides ancillary results on our experiments conducted on the MNIST datased. These results are similar to the results of the Shakespeare experiments.

\begin{figure*}[h]
\centering
\includegraphics[width=\textwidth]{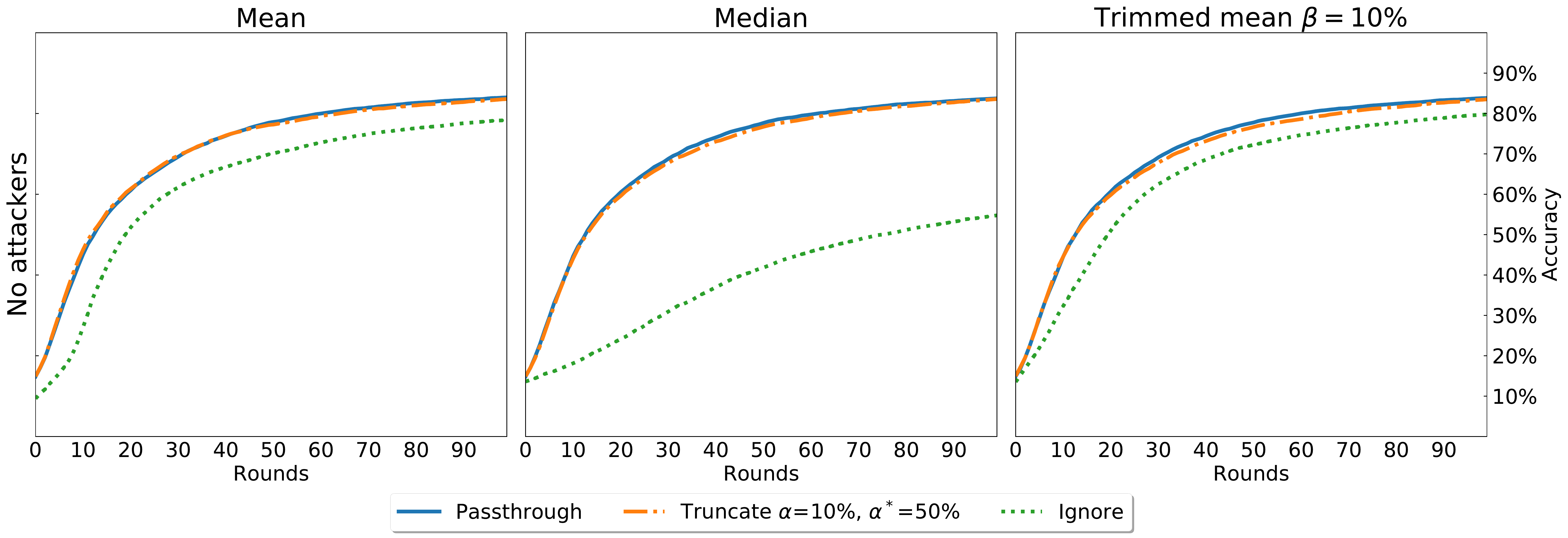}
\caption{Accuracy by round without any attackers for the MNIST experiments. Curves correspond to preprocessing procedures and columns correspond to different aggregation methods. It can be seen that our method (dashed orange curve) remains comparable to the properly weighted mean estimators (solid blue curve) while ignoring clients' sample sizes (dotted green curve) is sub-optimal. This effect is pronounced when the unweighted median is used, since with our unbalanced partition it is generally very far from the mean.}
\label{fig:mnist:no_attackers}
\end{figure*}

\begin{figure*}
\centering
\includegraphics[width=\textwidth]{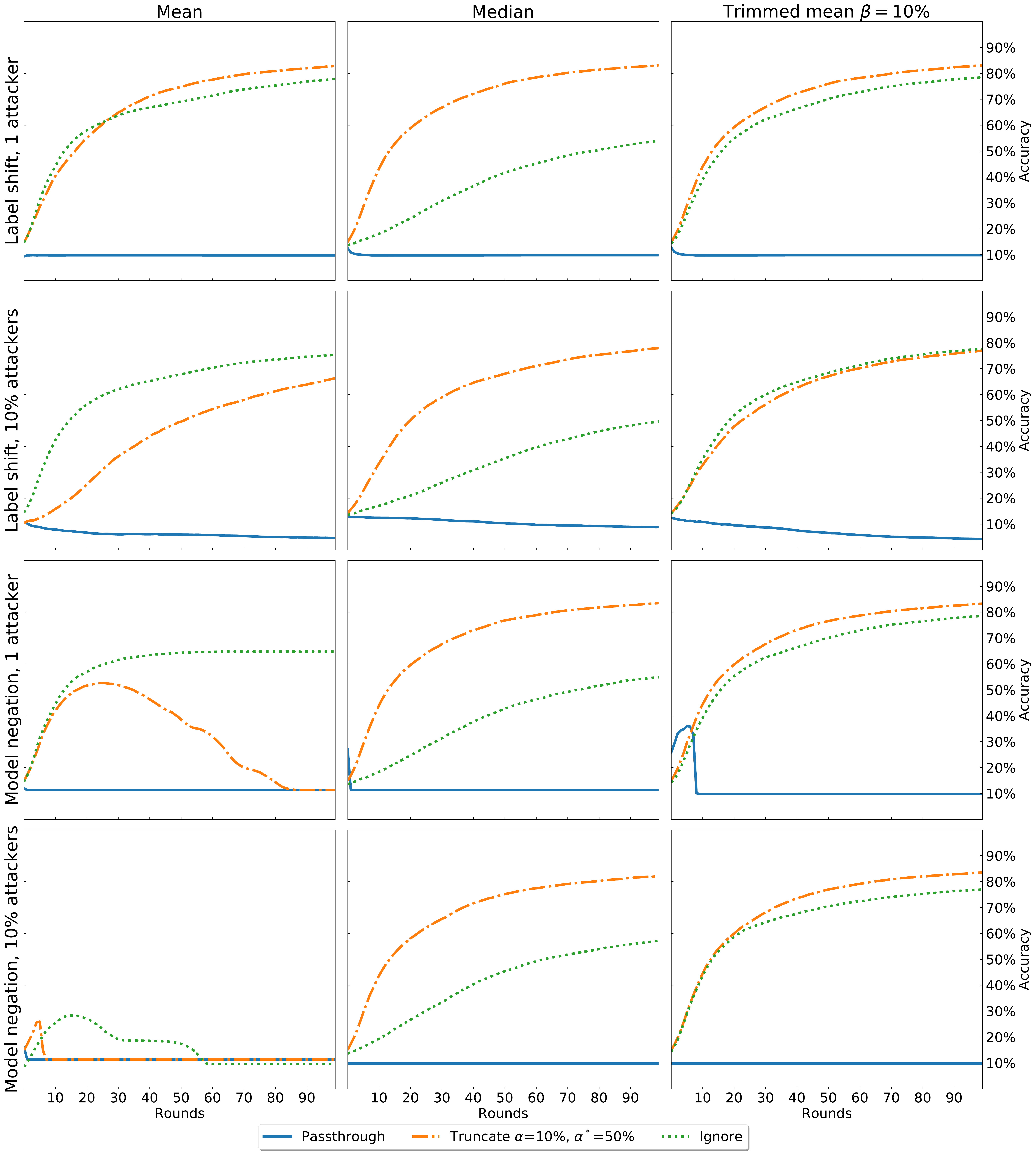}
\caption{Accuracy by round under Byzantine attacks for the MNIST experiments. Curves correspond to preprocessing procedures and columns correspond to different aggregation methods. In the first two rows Byzantine clients perform a label shifting attack with one and $10\%$ attackers, respectively. In the last two rows we repeat the experiment with a model negation attack.
\\\hspace{\textwidth}
We observe that even with a single attacker performing a trivial attack (first and third rows), using the weights directly (solid blue curve) is devastating while when our preprocessing method is used in conjunction with robust mean aggregations (dashed orange curve, two last columns) convergence remains stable even when there are actual $\alpha$ (=$10\%$) attackers (second and forth rows). We note that in some cases our method may be slightly less efficient compared with the preprocessing method that ignores sample size altogether (dotted green curve, second row, last column). This is to be expected because we allow Byzantine clients to potentially get close to $\alpha^*$-proportion ($50\%$, in this case) of the weight. However, our method is significantly closer to the optimal solution when there are no or only a few attackers (see Figure \ref{fig:mnist:no_attackers}). Moreover, when used in conjunction with robust mean aggregation methods it maintains their robustness properties.}
\label{fig:mnist:attack}
\end{figure*}

The experiment without any attackers is shown in Figure \ref{fig:mnist:no_attackers} and the executions with attackers are shown in Figure \ref{fig:mnist:attack}. 

% Additionally, final test accuracies are shown in Table \ref{final-acc}.
\vfill
\vfill

\end{document}